\documentclass[10pt,twocolumn,letterpaper]{article}

\usepackage[pagenumbers]{cvpr} 

\usepackage[T1]{fontenc}
\usepackage{graphicx}
\usepackage{amsmath,amssymb}
\usepackage{booktabs,multirow,array,tabularx}
\usepackage{xspace}
\usepackage{colortbl}
\usepackage{tikz}
\usepackage{marvosym}
\usetikzlibrary{arrows.meta,positioning,fit,calc}

\definecolor{tablefocus}{RGB}{231,241,244}
\definecolor{genericblue}{RGB}{225,237,250}
\definecolor{privategreen}{RGB}{228,242,230}
\definecolor{localorange}{RGB}{254,239,218}

\newcommand{\method}{CoPre-MoE\xspace}
\newcommand{\diag}{\operatorname{diag}}

\newcommand{\softmax}{\operatorname{softmax}}
\newcommand{\trunc}{\operatorname{SVD}}
\newcommand{\norm}[1]{\left\lVert #1\right\rVert}

\definecolor{cvprblue}{rgb}{0.21,0.49,0.74}
\usepackage[pagebackref,breaklinks,colorlinks,allcolors=cvprblue]{hyperref}

\def\confName{CVPR}
\def\confYear{2027}

\title{CoPre-MoE: Consolidating and Preserving MoE for \\Multimodal Continual Instruction Tuning}

\author{%
 {Tao Hu$^{1,2}$  \quad
Ling-Xiang Zhang$^{1,2}$    \quad
Da-Wei Zhou$^{1,2}\textsuperscript{(\Letter)}$} \\
$^{1} $ School of Artificial Intelligence, Nanjing University \\
$^{2} $ State Key Laboratory for Novel Software Technology, Nanjing University \\
\texttt{\small \{hut, zhanglx, zhoudw\}@lamda.nju.edu.cn}
}

\begin{document}
\maketitle

\begin{abstract}
Multimodal continual instruction tuning requires multimodal large language models to sequentially acquire new capabilities without catastrophic forgetting. 
Mixture-of-experts (MoE) architectures offer a promising way to mitigate task interference through expert specialization and combination. 
However, while existing MoE-based methods avoid overwriting shared parameters by allocating separate experts for each task, this progressive expansion incurs severe cross-task redundancy and linear parameter growth. 
Specifically, experts across tasks often learn overlapping functions alongside task-specific capabilities. For instance, ScienceQA and TextVQA both locate relevant image regions. 
To curb parameter redundancy without inducing forgetting, we propose Consolidating and Preserving MoE (\method).  
For each incoming task, \method first learns task-specific MoE to capture specialized capabilities. 
To eliminate redundancy and reduce parameter growth across tasks, we design a shared-and-private expert architecture.
Specifically, we first evaluate functional overlap to selectively consolidate overlapping experts across tasks into a fixed-capacity generic expert pool. We then factorize the remaining task-specific experts into shared projection bases and compact task-specific cores within a private expert pool. 
To mitigate forgetting, we construct the router by pairing each expert with learnable retrieval keys, where historical keys remain frozen during continual updates while newly added keys are regularized with an orthogonality penalty. 
At inference, input-dependent routing dynamically combines both pools without test-time task labels. 
Extensive experiments demonstrate that \method achieves superior performance while substantially reducing parameter overhead.
\end{abstract}

\begin{figure}[t]
  \centering
  \includegraphics[width=\linewidth]{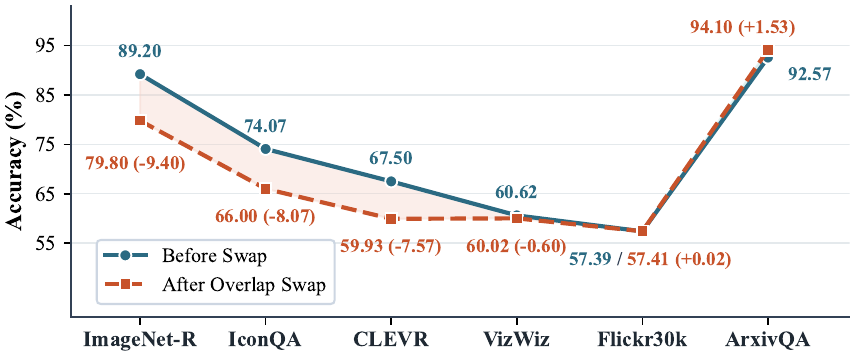}
\caption{\textbf{Empirical analysis of cross-task expert redundancy on UCIT.} 
Comparing native experts (\textbf{Before Swap}) with cross-task substitution (\textbf{After Overlap Swap}) shows that high-overlap experts preserve $90\%\sim 100\%$ of performance across tasks, confirming substantial functional redundancy while highlighting the need to preserve task-specific variations.}
  \label{fig:motivation_spectrum}
\end{figure}

\section{Introduction}
\label{sec:intro}
Multimodal large language models (MLLMs)~\cite{liu2024llava,dai2023instructblip,li2023blip,alayrac2022flamingo,bai2023qwenvl} acquire diverse capabilities through visual instruction tuning, including visual question answering, captioning, and complex reasoning. In practical deployment, models must continually learn new tasks and domain-specific knowledge from incoming streams. However, standard sequential fine-tuning suffers from catastrophic forgetting~\cite{mccloskey1989catastrophic,french1999catastrophic,kirkpatrick2017overcoming,delange2022survey}, where newly learned representations overwrite parameters critical to historical tasks. Multimodal continual instruction tuning (MCIT)~\cite{chen2024coin,guo2025hide,chen2025sefe,zeng2025modalprompt,hu2026drape} formalizes this challenge, requiring models to sequentially master incoming instruction streams while preserving earlier capabilities without relying on data replay. Mixture-of-experts (MoE) architectures~\cite{shazeer2017moe,fedus2022switch,wu2024mole,dou2023loramoe} offer a promising solution for MCIT by assigning modular network capacities to specialized transformations and conditionally composing them via input-dependent routing~\cite{wang2025smolora,xie2026same,ge2025curriculum,huai2025clmoe}.

However, existing MoE-based MCIT methods struggle to balance parameter efficiency with knowledge preservation. To avoid overwriting historical capabilities, modular expansion and decoupling approaches~\cite{wang2025smolora,guo2025hide,xie2026same,huai2025clmoe} allocate separate experts or constrain routing for each incoming task. Yet, as tasks accumulate, total parameter storage grows linearly while distinct experts repeatedly acquire overlapping functions across the stream. Conversely, strategies that directly update shared experts or average weights across tasks~\cite{chen2024coin,chen2025sefe,ilharco2023task,yadav2023ties} restrict parameter growth, but forcing incompatible task adaptations into shared parameters corrupts historical capabilities and induces severe forgetting. Consequently, existing frameworks fail to resolve this fundamental dilemma between cross-task redundancy and parameter overwriting.

To empirically investigate this cross-task functional redundancy, we sequentially train a single MLLM equipped with task-specific MoE adapters across six vision-language tasks on UCIT~\cite{guo2025hide} (Figure~\ref{fig:motivation_spectrum}). After sequential training, we compare two inference settings for each task: (i)~\emph{Before Swap}, which retains all native experts learned on the current task; and (ii)~\emph{After Overlap Swap}, which directly replaces the high-overlap expert subset in the current task with the most functionally similar experts from other tasks based on minimal activation-weighted output distortion.\footnote{The exact formulation of activation-weighted functional distortion is detailed in Sec.~\ref{sec:merge}.} We then measure the resulting task scores across all six tasks to examine how cross-task expert substitution affects performance.

\looseness=-1
Figure~\ref{fig:motivation_spectrum} compares task performance across the sequential stream before and after cross-task expert substitution. Specifically, substituting a native expert with its most functionally similar cross-task counterpart maintains near-identical accuracy on tasks like VizWiz, Flickr30k, and ArxivQA (e.g., $57.39 \rightarrow 57.41$ on Flickr30k), while distinct tasks like ImageNet-R, IconQA, and CLEVR still retain roughly 90\% of their original performance (e.g., achieving $79.80$ on ImageNet-R). In summary, these empirical observations yield two key insights. (1)~Learned experts across tasks consistently acquire shared functional transformations, confirming that allocating separate experts per task incurs substantial cross-task redundancy. (2)~Meanwhile, the modest performance gap observed under naive replacement indicates that directly discarding native experts inevitably loses their learned capabilities, demonstrating the need for an effective consolidation strategy that merges redundant capabilities while preserving information from both participating experts.

Guided by this insight, we propose Consolidating and Preserving MoE (\method), a selective dual-pool framework for MCIT. For each incoming task, \method first learns multiple low-rank experts along with a router on a frozen MLLM backbone. To curb redundancy and reduce parameter growth, we design a shared-and-private expert organization based on input activation statistics accumulated during training. Specifically, we first evaluate functional overlap to selectively consolidate high-overlap experts across tasks into a fixed-capacity \emph{generic expert pool} via activation-weighted operator merging, eliminating cross-task redundancy while preserving critical feature channels. We then factorize the remaining task-specific experts into shared projection bases and compact task-specific cores within a \emph{private expert pool}, preserving specialized capabilities at a fraction of the parameter cost. Furthermore, to mitigate forgetting, we construct the router by pairing each expert with learnable retrieval keys, where historical keys remain frozen during continual updates while newly added keys are regularized to be orthogonal to past ones. This enables input-dependent routing to dynamically query and combine both pools at inference without task labels. Extensive evaluations on UCIT and CoIN demonstrate that \method achieves state-of-the-art performance while reducing adaptation parameter overhead by over 68\% compared with MoELoRA.

\section{Related Work}
\label{sec:related}
\noindent\textbf{Mixture-of-experts and parameter-efficient adaptation.}
Mixture-of-experts (MoE) architectures scale effective model capacity without a proportional increase in computational cost by activating conditional subsets of network parameters via input-dependent gating~\cite{shazeer2017moe,fedus2022switch}. In parallel, parameter-efficient fine-tuning (PEFT) adapts frozen foundation models by introducing lightweight modular updates, such as low-rank matrices or residual bottleneck adapters~\cite{hu2022lora,zhang2023llamaadapter,wang2023olora}. Recent works combine these paradigms by deploying low-rank adapters as modular experts. For instance, MoLE~\cite{wu2024mole} learns hierarchical input-dependent weights over trained LoRA branches, while LoRA-MoE~\cite{dou2023loramoe} trains specialized low-rank modules to process diverse instruction distributions. These frameworks predominantly focus on inference-time expert activation and gating design over pre-allocated modular branches.

\noindent\textbf{Multimodal Continual Instruction Tuning.}
Continual learning investigates learning from non-stationary data streams without catastrophic forgetting~\cite{delange2022survey}, employing strategies such as parameter isolation~\cite{mallya2018packnet}, weight regularization~\cite{kirkpatrick2017overcoming}, and prompt pooling~\cite{wang2022learning,wang2022dualprompt}. In vision-language domains, CoIN~\cite{chen2024coin} formalized multimodal continual instruction tuning (MCIT) across heterogeneous capabilities. Subsequent research broadly explores prompt-based adaptation and modular expansion. Prompt-based approaches prepend visual or instruction-guided tokens to steer pre-trained representations~\cite{zeng2025modalprompt,hu2026drape,wang2024m2pt,wang2024inscl}. Modular approaches allocate dedicated low-rank adapters or MoE components to incoming tasks~\cite{wang2025smolora,ge2025curriculum,huai2025clmoe}. To mitigate cross-task interference, recent methods introduce decoupled representations, subspace constraints, or routing stabilization: HiDe-LLaVA~\cite{guo2025hide} separates task-generic and task-specific parameters, SEFE~\cite{chen2025sefe} preserves essential parameter subsets, KeepLoRA~\cite{luo2026keeplora} constrains low-rank updates within residual gradient subspaces, and SAME~\cite{xie2026same} stabilizes gating distributions across stages.

\section{Preliminaries}
\label{sec:preliminaries}
\subsection{Multimodal Continual Instruction Tuning}
\label{sec:prelim-mcit}
MCIT considers a sequence of $T$ tasks. At stage $t\in\{0,\ldots,T-1\}$, the model receives a task-specific instruction dataset $\mathcal{D}_t=\{(v_n^t,I_n^t,y_n^t)\}_{n=1}^{n_t}$, where $v$, $I$, and $y$ denote an image, a textual instruction, and its target response, respectively. Given a pre-trained MLLM backbone $\theta$ and trainable adaptation parameters $\psi$, learning on stage $t$ minimizes the autoregressive next-token prediction loss:
\begin{equation}
 \mathcal{L}_t(\psi)=-\mathbb{E}_{(v,I,y)\sim\mathcal{D}_t}
 \sum_{\nu=1}^{|y|}\log p_{\theta,\psi}(y_\nu\mid v,I,y_{<\nu}).
 \label{eq:prelim-nll}
\end{equation}
\looseness=-1
During this continual learning process, raw examples from earlier tasks are unavailable, and the model only retains learned adaptation parameters and compact statistics. After each stage $t$, it is evaluated on all seen tasks to measure acquisition and retention~\cite{chen2024coin,guo2025hide}. Task indices define the training sequence, but are not supplied at inference.

\subsection{Mixture-of-Experts Adaptation}
\label{sec:prelim-moe}
An MoE layer conditionally routes modular expert transformations through an input-dependent gating mechanism~\cite{shazeer2017moe,fedus2022switch}. For parameter-efficient adaptation, each expert is parameterized as a low-rank residual update $D_i=sB_iA_i$ alongside a frozen linear weight matrix $W$~\cite{hu2022lora,wu2024mole}, where $A_i\in\mathbb R^{r\times d_{\rm in}}$, $B_i\in\mathbb R^{d_{\rm out}\times r}$, $r$ is the inner rank, and $s$ is a scaling factor. For an intermediate feature input $x$, the adapted module output at stage $t$ is formulated as:
\begin{equation}
 y=Wx+\sum_{i\in\mathcal{E}_t}\rho_i(\mathbf{q})D_i x,
 \label{eq:prelim-experts}
\end{equation}
where $\mathcal{E}_t$ denotes the set of active experts available up to stage $t$, and $\rho_i(\mathbf{q})$ represents the input-dependent routing weight derived from the multimodal instruction query $\mathbf{q}$, with $\sum_{i\in\mathcal{E}_t}\rho_i(\mathbf{q})=1$. In continual learning, modular MoE approaches prevent catastrophic forgetting by expanding $\mathcal{E}_t$ with newly allocated experts for each incoming task while keeping historical experts frozen~\cite{wang2025smolora,ge2025curriculum}. However, independently storing all $|\mathcal{E}_t|$ rank-$r$ experts requires $|\mathcal{E}_t|r(d_{\rm in}+d_{\rm out})$ factor parameters per module, causing total storage to scale linearly with the number of tasks.

\noindent\textbf{Discussion.} While expanding $\mathcal{E}_t$ in Equation~\ref{eq:prelim-experts} prevents historical parameter overwriting, this linear expansion introduces severe cross-task functional redundancy. Specifically, because experts in $\mathcal{E}_t$ are trained independently on sequential tasks without cross-task coordination, multiple experts across different tasks repeatedly learn overlapping functional transformations ($D_i x \approx D_j x$, as demonstrated in Figure~\ref{fig:motivation_spectrum}), alongside task-unique adaptations. Consequently, existing expansion-based MoE methods that indiscriminately preserve all per-task experts in $\mathcal{E}_t$ waste substantial parameter capacity on redundant functions. To overcome this limitation, an effective continual MoE framework should selectively consolidate high-overlap experts across tasks to eliminate redundancy, while preserving and compressing the remaining task-specific experts to maintain specialized capabilities with compact parameter overhead.

\begin{figure*}[t]
\centering
\includegraphics[width=\linewidth]{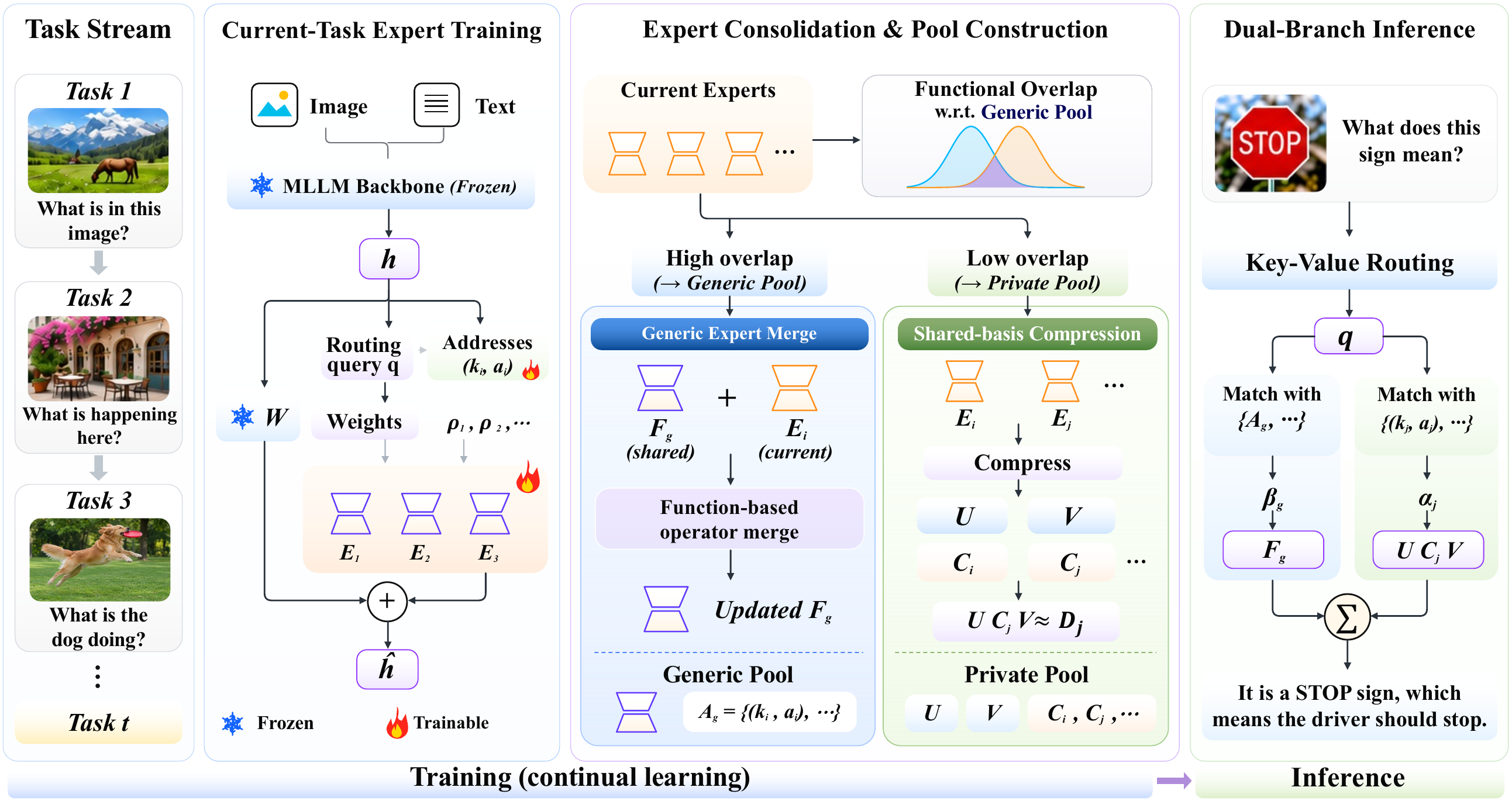}
\vspace{-6mm}
\caption{\textbf{Overview of \method.}
\textbf{Left:} Task-specific low-rank experts and key-value routing pairs are trained on a frozen MLLM backbone.
\textbf{Middle:} At task boundaries, functionally overlapping experts are consolidated into a fixed-capacity Generic Pool, while task-specific experts are factorized into shared bases and compact cores in a Private Pool.
\textbf{Right:} Routing query $\mathbf{q}$ dynamically queries both pools to compose generic and private operators for inference.}
\label{fig:overview}
\end{figure*}

\section{Method}
\label{sec:method}
To address the cross-task redundancy and parameter growth inherent in continual MoE expansion, \method establishes a selective dual-pool architecture with an orthogonal key-value router. It consolidates overlapping expert transformations into a fixed-capacity generic pool while factorizing task-specific adaptations into a compact private pool.

\subsection{Learning Routed Low-Rank Experts}
\label{sec:setup}
\noindent\textbf{Expert parameterization and training.}
Following Sec.~\ref{sec:prelim-mcit}, \method freezes the pre-trained vision tower, language model, and multimodal projector. For each adapted linear module $\ell$, an incoming task $t$ trains a task-specific MoE comprising $M$ rank-$r$ experts:
\begin{equation}
 D_{t,i}^{\ell}=s B_{t,i}^{\ell} A_{t,i}^{\ell},\qquad i\in\{0,\ldots,M-1\},
 \label{eq:lora}
\end{equation}
where $A_{t,i}^{\ell}\in\mathbb{R}^{r\times d_{\rm in}}$ and $B_{t,i}^{\ell}\in\mathbb{R}^{d_{\rm out}\times r}$ are low-rank factor matrices. To support input-dependent retrieval, we construct the router by pairing each expert $i$ with a learnable retrieval key $k_i\in\mathbb{R}^d$ and a feature attention vector $a_i\in\mathbb{R}^d$, forming a persistent routing pair $(k_i, a_i)$ matched against a routing query $\mathbf q\in\mathbb{R}^d$. Immediately before response generation, $\mathbf q$ is extracted from a frozen shallow language layer at the instruction-response boundary and reused across subsequent response tokens. The routing similarity $c_i(\mathbf q)$ and logit $l_i(\mathbf q)$ for expert $i$ are computed as:
\begin{equation}
 c_i(\mathbf q)=\cos(\mathbf q\odot a_i,k_i),\qquad l_i(\mathbf q)=c_i(\mathbf q)/\tau,
 \label{eq:address}
\end{equation}
where $\odot$ denotes element-wise multiplication and $\tau$ is a temperature hyperparameter. Within task $t$, the routing logits are normalized via $\rho_{t,i}(\mathbf q)=\softmax_i(l_i(\mathbf q))$ to gate the $M$ experts and produce the adapted module output:
\begin{equation}
 y_{t}^{\ell}=W^{\ell}x^{\ell}
 +\sum_{i=0}^{M-1}\rho_{t,i}(\mathbf q)D_{t,i}^{\ell}x^{\ell}.
 \label{eq:train-forward}
\end{equation}
Throughout continual adaptation, all historical expert parameters and their routing pairs remain frozen to prevent catastrophic forgetting. To avoid routing collisions between new and old tasks, training on stage $t>0$ optimizes the next-token prediction loss $\mathcal{L}_{\rm NLL}=\mathcal{L}_t(\psi)$ (Eq.~\ref{eq:prelim-nll}) alongside an orthogonality penalty between current and historical routing pairs:
\begin{equation}
 \begin{split}
 \mathcal{L}={}&\mathcal{L}_{\rm NLL}+\lambda_{\rm orth}\big[
 \operatorname{mean}\big((K_tK_{<t}^{\top})^{\odot2}\big)\\
 &+\operatorname{mean}\big((\mathsf A_t\mathsf A_{<t}^{\top})^{\odot2}\big)\big],
 \end{split}
 \label{eq:loss}
\end{equation}
where $K_t,\mathsf A_t$ and $K_{<t},\mathsf A_{<t}$ denote row-normalized current and historical key and attention matrices, respectively, and $\lambda_{\rm orth}$ weights the penalty.

Upon completing stage $t$, the model obtains a newly learned expert set $\mathcal{E}_t^{\rm new}=\{D_{t,i}^{\ell}\}_{i=0}^{M-1}$ and its corresponding routing pairs $\{(k_{t,i}, a_{t,i})\}_{i=0}^{M-1}$. As analyzed in Sec.~\ref{sec:prelim-moe}, directly accumulating $\mathcal{E}_t^{\rm new}$ across tasks causes severe cross-task functional redundancy and linear parameter growth. We therefore partition $\mathcal{E}_t^{\rm new}$ at each task boundary, selectively consolidating high-overlap experts into a shared generic pool while factorizing the remaining task-specific experts into a private pool.

\subsection{Selective Generic Consolidation}
\label{sec:merge}
Maintaining a fixed-capacity generic pool requires deciding both which incoming experts in $\mathcal{E}_t^{\rm new}$ to consolidate and where to place them. The generic pool has a fixed capacity of $G$ slots, each storing a rank-$r$ operator $F_g^{\ell}$ and a set $\mathcal A_g$ of source routing pairs. We evaluate candidate assignments by the functional distortion they introduce to represented operators, using compact activation summaries accumulated during training.

\noindent\textbf{Function-based operator merge.}
To consolidate an incoming expert with an occupied generic slot, we estimate their functional compatibility from retained activation summaries. For each current expert $i$ and module $\ell$, \method tracks diagonal second-moment statistics during training:
\begin{equation}
 s_{t,i}^{\ell}=\sum_b\rho_{b,i}\left(\frac{1}{L_b}
 \sum_{n=1}^{L_b}x_{b,n}^{\ell}\odot x_{b,n}^{\ell}\right).
 \label{eq:stats}
\end{equation}
Here $b$ indexes training sequences of length $L_b$, $x_{b,n}^\ell$ is the module input at token position $n$, and $\rho_{b,i} = \rho_{t,i}(\mathbf{q}_b)$ is the sequence routing weight. Activations are averaged over tokens before routing-weighted accumulation. When sources are merged into a generic slot, their second moments are summed and retained. For a candidate consolidation pair $\mathcal{S}=\{g, i\}$ combining generic slot $g$ with incoming expert $i$, coordinate-wise weights $w_m^\ell$ form the optimal target operator $\overline{D}^\ell$ that minimizes expected squared output deviation:
\begin{equation}
 w_m^{\ell}=\frac{s_m^{\ell}}{\sum_{m'\in\mathcal{S}}s_{m'}^{\ell}},\qquad
 \overline{D}^{\ell}=\sum_{m\in\mathcal{S}}D_m^{\ell}\diag(w_m^{\ell}),
 \label{eq:target}
\end{equation}
where $D_g^\ell = F_g^\ell$ and $s_g^\ell$ denote the existing slot's operator and accumulated second moments, while $D_i^\ell$ and $s_i^\ell$ denote those of incoming expert $i$. Divisions in Eq.~\ref{eq:target} are coordinate-wise with zero-energy protection. To preserve the original expert rank budget $r$, we project the merged operator back to rank $r$ via truncated SVD as $\widehat{D}^{\ell}=\trunc_r(\overline{D}^{\ell})$, retaining the leading $r$ singular components. Compact QR factorizations reduce this projection to a small matrix, avoiding full-matrix materialization. To quantify functional overlap, we score candidate $\widehat{D}^\ell$ against each source $m\in\mathcal{S}$ by its relative functional distortion:
\begin{equation}
 e_m=\frac{\sum_{\ell}\norm{(\widehat{D}^{\ell}-D_m^{\ell})
 \diag(s_m^{\ell})^{1/2}}_F^2}
 {\max\left(\sum_{\ell}\norm{D_m^{\ell}\diag(s_m^{\ell})^{1/2}}_F^2,\epsilon\right)},
 \label{eq:error}
\end{equation}
with Frobenius norm $\norm{\cdot}_F$ and $\epsilon>0$ for stability.

\noindent\textbf{History-weighted boundary search.}
At each task boundary, \method jointly selects a subset of high-overlap incoming experts $\mathcal{E}_{\rm in}\subset\mathcal{E}_t^{\rm new}$ and distinct target generic slots $\mathcal{M}$ ($|\mathcal{M}|=|\mathcal{E}_{\rm in}|$) to minimize:
\begin{equation}
 J=\frac{\sum_{g\in\mathcal{M}}N_g e_g+\sum_{i\in\mathcal{E}_{\rm in}}e_i}
 {\sum_{g=0}^{G-1}N_g+|\mathcal{E}_{\rm in}|},
 \label{eq:history}
\end{equation}
where $e_g, e_i$ denote source distortions (Eq.~\ref{eq:error}) and $N_g$ counts historical sources represented by slot $g$. Untouched slots contribute zero additional distortion. History weighting penalizes altering slots with more represented sources, while distinct assignment prevents multiple incoming experts from entering the same slot. Selected slots update their operators $F_g^{\ell}\leftarrow\widehat{D}^{\ell}$, increment source counts $N_g\leftarrow N_g+1$, and append each source's routing pair to $\mathcal{A}_g$, thereby consolidating redundant cross-task functions into fixed-capacity generic operators $\{F_g^{\ell}\}_{g=0}^{G-1}$.

\subsection{Compact Private Preservation}
\label{sec:pca}
\label{sec:complexity}
\noindent\textbf{Subspace compression with shared bases.}
Remaining incoming experts $\mathcal{E}_t^{\rm new}\setminus\mathcal{E}_{\rm in}$ outside the generic selection capture task-specific adaptations and are routed to the private pool, where each component retains its individual routing pair. For the compression derivation below, the module index $\ell$ is omitted. We represent private components via a compact Tucker-style decomposition~\cite{tucker1966notes}. Let $P_j\in\mathbb{R}^{d_{\rm out}\times d_{\rm in}}$ denote each private component matrix, including both historical approximations $\widehat{P}_j$ and unmerged incoming experts $D_{t,i}$. Orthonormal QR bases $Q_U,Q_V$ span $\{P_j\}$ in a compact subspace. We project each private operator $P_j$ and normalize its energy:
\begin{equation}
 H_j=Q_U^{\top}P_jQ_V,\qquad \widetilde{H}_j=H_j/\max(\norm{H_j}_F,\epsilon).
 \label{eq:compact}
\end{equation}
Energy normalization prevents high-magnitude components from dominating subspace selection. Retaining the leading $R$ eigenvectors $E_U,E_V$ of the two scatter matrices:
\begin{equation}
 S_U=\sum_j\widetilde{H}_j\widetilde{H}_j^{\top},\qquad
 S_V=\sum_j\widetilde{H}_j^{\top}\widetilde{H}_j,
 \label{eq:pca}
\end{equation}
yields updated shared bases $U, V$ and task-specific compact cores $C_j\in\mathbb{R}^{R\times R}$:
\begin{equation}
 U=Q_UE_U,\quad V=E_V^{\top}Q_V^{\top},\quad
 C_j=E_U^{\top}H_jE_V.
 \label{eq:pcaout}
\end{equation}
Cores are projected from unnormalized $H_j$, preserving component amplitudes. Each private expert is thus preserved as $\widehat P_j^{\ell}=U^{\ell}C_j^{\ell}V^{\ell}$, where $U^{\ell}\in\mathbb R^{d_{\rm out}\times R}$ and $V^{\ell}\in\mathbb R^{R\times d_{\rm in}}$ are shared across the pool and $C_j^{\ell}\in\mathbb R^{R\times R}$ is a task-specific compact core. While individual experts have rank $r$, the shared bases span a richer $R$-dimensional subspace ($R>r$). Each core absorbs the original LoRA scaling once. Original factors are discarded after compression, so subsequent boundaries compress already approximated functions while preserving component identity.

\subsection{Dual-Branch Inference and Routing}
\label{sec:routing}
\noindent\textbf{Generic and private pool routing.}
In the generic pool, each consolidated operator $F_g^{\ell}$ retains the set of source routing pairs $\mathcal{A}_g=\{(k_j, a_j)\}$. Rather than averaging keys, the routing logit for slot $g$ aggregates individual similarities $c_j=c_j(\mathbf q)$ (Eq.~\ref{eq:address}) via log-mean-exp:
\begin{equation*}
 z_g=\tau_a\log(\frac{1}{|\mathcal{A}_g|}
 \sum_{j\in\mathcal{A}_g}\exp(c_j/\tau_a)),
 \beta_g=\softmax_g(z_g/\tau),
\end{equation*}
where $\tau_a$ denotes the aggregation temperature and $\beta_g$ normalizes over all $G$ generic slots, preventing slots with more accumulated sources from overly dominating softmax probabilities. In parallel, each private component $j$ from task $u(j)$ is routed via its individual routing logit $l_j=l_j(\mathbf q)$ (Eq.~\ref{eq:address}), normalized across all private components $m$:
\begin{equation}
 \alpha_j=\frac{\pi_{u(j)}\exp(l_j)}{\sum_m\pi_{u(m)}\exp(l_m)},
 \label{eq:privateroute}
\end{equation}
where $\pi_u=\softmax_u\big((\lambda_v\cos(f_v,p_u^v)+\lambda_I\cos(f_I,p_u^I))/\tau_{p}\big)$ serves as a smooth domain-level prior computed from frozen visual-textual input features $(f_v,f_I)$ and running mean task centroids $(p_u^v,p_u^I)$ with balancing weights $\lambda_v, \lambda_I$ and temperature $\tau_{p}$.

\noindent\textbf{Summary.}
Combining generic consolidation and private compression, \method replaces linear MoE expansion with a compact dual-pool inference pipeline. Given an input $(v,I)$ without task labels, the router extracts query $\mathbf{q}$ to compute generic weights $\{\beta_g\}$ and private weights $\{\alpha_j\}$ in parallel. The adapted forward pass combines the frozen backbone with both pools:
\[
y^{\ell}=
\underbrace{W^{\ell}x^{\ell}\vphantom{U^{\ell}\left(\sum_j\alpha_j C_j^{\ell}\right)V^{\ell}x^{\ell}}}_{\text{Frozen Backbone}}
+\textcolor[HTML]{547A9E}{\underbrace{\sum_{g=0}^{G-1}\beta_g F_g^{\ell}x^{\ell}\vphantom{U^{\ell}\left(\sum_j\alpha_j C_j^{\ell}\right)V^{\ell}x^{\ell}}}_{\text{Generic Pool}}}
+\textcolor[HTML]{B87970}{\underbrace{U^{\ell}\left(\sum_j\alpha_j C_j^{\ell}\right)V^{\ell}x^{\ell}}_{\text{Private Pool}}}.
\]
Here, the generic term activates consolidated cross-task knowledge without storing redundant expert copies, while the private term reconstructs task-specific adaptations. Because the compact core mixture $\sum_j \alpha_j C_j^{\ell}$ is evaluated first, each module requires only a single application of the shared projections $V^{\ell}$ and $U^{\ell}$, ensuring low inference latency. Once the generic pool is initialized, its frozen branch $\sum_{g=0}^{G-1}\beta_g F_g^{\ell}x^{\ell}$ is attached to Eq.~\ref{eq:train-forward} during subsequent task training to facilitate forward knowledge transfer.

\begin{table*}[t]
\centering
\caption{Final task accuracy and backward transfer on UCIT ($\uparrow$: higher is better). Bold and underline denote best and second-best results.}
\label{tab:ucit-comparison}
\small
\setlength{\tabcolsep}{5.5pt}
\begin{tabular}{lcccccccc}
\toprule
Method & ImageNet-R & ArxivQA & VizWiz & IconQA & CLEVR & Flickr30k & $\overline{\mathcal{A}}\uparrow$ & BWT$\uparrow$\\
\midrule
Zero-shot~\cite{liu2024llava} &16.27&53.73&38.39&19.20&20.63&41.88&31.68&--\\
LoRA-FT~\cite{hu2022lora} &58.03&77.63&44.39&67.40&61.77&\textbf{58.22}&61.24&-18.78\\
MoELoRA~\cite{chen2024coin} &70.07&77.70&44.69&50.03&54.03&57.34&58.98&-14.63\\
SEFE~\cite{chen2025sefe} &80.83&78.00&47.01&\underline{69.63}&\underline{65.83}&\underline{57.92}&66.54&-11.33\\
HiDe-LLaVA~\cite{guo2025hide} &\underline{84.03}&90.73&44.43&58.93&41.37&54.25&62.29&-9.20\\
SAME~\cite{xie2026same} &83.83&\underline{91.40}&\underline{51.33}&65.27&53.50&57.43&\underline{67.12}&\underline{-2.41}\\
\midrule
\rowcolor{tablefocus}
\textbf{\method} &\textbf{88.93}&\textbf{93.73}&\textbf{62.83}&\textbf{74.07}&\textbf{66.63}&57.39&\textbf{73.93}&\textbf{-0.10}\\
\bottomrule
\end{tabular}
\end{table*}

\begin{table*}[t]
\centering
\caption{Final task accuracy and backward transfer on CoIN ($\uparrow$: higher is better). Bold and underline denote best and second-best results.}
\label{tab:coin-comparison}
\small
\setlength{\tabcolsep}{4.2pt}
\begin{tabular}{lcccccccccc}
\toprule
Method & ScienceQA & TextVQA & ImageNet & GQA & VizWiz & Grounding & VQAv2 & OCR-VQA & $\overline{\mathcal{A}}\uparrow$ & BWT$\uparrow$\\
\midrule
Zero-shot~\cite{liu2024llava} &67.92&57.71&47.37&61.28&45.17&6.12&52.03&53.58&48.89&--\\
LoRA-FT~\cite{hu2022lora} &26.00&25.38&28.51&33.07&26.52&0.10&40.00&52.92&29.06&-31.16\\
MoELoRA~\cite{chen2024coin} &62.02&52.05&37.21&53.12&43.32&33.22&57.92&\underline{65.75}&50.58&-19.04\\
SEFE~\cite{chen2025sefe} &75.35&58.66&83.10&54.25&48.85&16.75&65.35&\textbf{66.25}&58.57&-11.94\\
HiDe-LLaVA~\cite{guo2025hide} &73.20&56.92&69.28&\underline{61.33}&50.76&59.18&\textbf{67.12}&64.76&62.82&-6.38\\
SAME~\cite{xie2026same} &\underline{78.35}&\underline{60.69}&\underline{90.21}&\textbf{61.70}&\underline{54.13}&\underline{59.87}&66.04&63.59&\underline{66.82}&\underline{-4.23}\\
\midrule
\rowcolor{tablefocus}
\textbf{\method} &\textbf{81.09}&\textbf{62.37}&\textbf{97.03}&61.13&\textbf{56.75}&\textbf{70.87}&\underline{66.60}&64.12&\textbf{70.00}&\textbf{0.02}\\
\bottomrule
\end{tabular}
\end{table*}

\section{Experiments}
\label{sec:experiments}

\begin{table*}[t]
\centering
\caption{UCIT ablation study evaluating expert allocation, function consolidation, and routing mechanisms. Higher $\overline{\mathcal{A}}$ and BWT are better ($\uparrow$).}
\label{tab:module-ablation}
\small
\setlength{\tabcolsep}{4.0pt}
\begin{tabular}{lcccccccc}
\toprule
Setting & ImageNet-R & ArxivQA & VizWiz & IconQA & CLEVR & Flickr30k & $\overline{\mathcal{A}}\uparrow$ & BWT$\uparrow$\\
\midrule
\rowcolor{tablefocus}
\method (Full model) &{88.93}&{93.73}&{62.83}&{74.07}&{66.63}&{57.39}&{73.93}&{-0.10}\\
w/o Dynamic role allocation &88.17&91.27&60.66&73.07&64.80&56.54&72.42&-0.52\\
w/o Activation-weighted merge &88.93&94.10&62.22&73.20&61.20&56.96&72.77&-0.02\\
w/o Multi-key retention &88.87&93.23&62.12&73.93&62.17&57.38&72.95&-0.45\\
\bottomrule
\end{tabular}
\end{table*}
\subsection{Implementation Details}
\label{sec:protocol}
\noindent\textbf{Datasets and baselines.}
We evaluate on two sequential benchmarks following prior MCIT literature~\cite{chen2024coin,guo2025hide,hu2026drape}. UCIT~\cite{guo2025hide} includes ImageNet-R~\cite{hendrycks2021many}, ArxivQA~\cite{li2024multimodal}, VizWiz~\cite{gurari2018vizwiz}, IconQA~\cite{lu2021iconqa}, CLEVR~\cite{lindstrom2022clevr}, and Flickr30k~\cite{plummer2015flickr30k}. CoIN~\cite{chen2024coin} covers ScienceQA~\cite{lu2022learn}, TextVQA~\cite{singh2019towards}, ImageNet~\cite{deng2009imagenet}, GQA~\cite{hudson2019gqa}, VizWiz~\cite{gurari2018vizwiz}, Grounding (RefCOCO)~\cite{kazemzadeh2014referitgame}, VQAv2~\cite{goyal2017making}, and OCR-VQA~\cite{mishra2019ocr}. Both benchmarks follow standard acquisition sequences and evaluation protocols. We compare \method against Zero-shot~\cite{liu2024llava} alongside five representative continual baselines: LoRA-FT~\cite{hu2022lora}, MoELoRA~\cite{chen2024coin}, SEFE~\cite{chen2025sefe}, HiDe-LLaVA~\cite{guo2025hide}, and SAME~\cite{xie2026same}.

\noindent\textbf{Training setup.}
We use LLaVA-1.5-7B with CLIP ViT-L/14-336~\cite{liu2024llava,radford2021clip} in Prism~\cite{tang2026prism}. Low-rank experts adapt all attention and FFN linear projections across layers 1--31 with rank $r=4$ and scaling $s=2$. Each task trains $M=4$ experts for one epoch with learning rate $10^{-4}$ and global batch size 48 on RTX 4090 GPU. Generic capacity is $G=4$, initialized after Task~1 and consolidating $|\mathcal{E}_{\rm in}|=2$ experts per boundary. Default private rank is $R=16$. Routing weights use $\lambda_v=\lambda_I=0.5$ and orthogonality penalty $\lambda_{\rm orth}=1.0$. Training uses BF16, ZeRO-2, and cosine decay, while backbones remain frozen.

\noindent\textbf{Evaluation metrics.}
Let $R_{t,i}$ denote the evaluation accuracy on task $i$ after training on stage $t$. Following standard continual learning protocols~\cite{lopez2017gem,chen2024coin}, we report the final average accuracy across all $T$ tasks $\overline{\mathcal{A}} = \frac{1}{T}\sum_{i=0}^{T-1} R_{T-1,i}$ and backward transfer $\operatorname{BWT} = \frac{1}{T-1}\sum_{i=0}^{T-2}(R_{T-1,i}-R_{i,i})$. BWT is inapplicable to Zero-shot.

\subsection{Benchmark Comparison}
\label{sec:comparison}
Across both sequential benchmarks, \method establishes new state-of-the-art retention and overall performance without test-time task labels. On UCIT (Table~\ref{tab:ucit-comparison}), \method achieves a final average score of $\overline{\mathcal{A}}=73.93$ and near-zero backward transfer ($\text{BWT}=-0.10$), outperforming the strongest baseline SAME ($67.12$) by $6.81$ points, with notable gains on challenging benchmarks such as CLEVR ($+13.13$) and VizWiz ($+11.50$). This superiority consistently generalizes to the longer eight-task sequence on CoIN (Table~\ref{tab:coin-comparison}), where \method achieves $\overline{\mathcal{A}}=70.00$ and $\text{BWT}=0.02$, exceeding SAME by $3.18$ points with substantial improvements on Grounding ($+11.00$), ScienceQA ($+2.74$), and TextVQA ($+1.68$). These results demonstrate that selectively consolidating shared generic transformations while isolating private low-rank subspaces effectively alleviates catastrophic forgetting across continual multimodal instruction streams.

\begin{figure}[t]
\centering
\includegraphics[width=\linewidth]{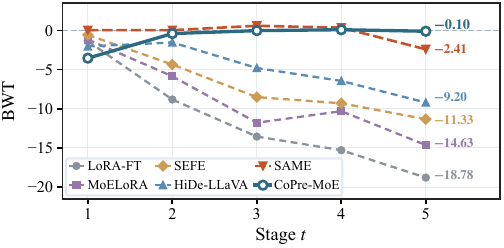}
\vspace{-2mm}
\caption{Backward transfer trajectory across sequential training stages $t$ on UCIT. Points evaluate average BWT over previously learned tasks; terminal values report final retention.}
\label{fig:retention-trajectory}
\end{figure}

\noindent\textbf{Retention dynamics across stages.}
Figure~\ref{fig:retention-trajectory} illustrates backward transfer trajectories across sequential training stages on UCIT. While LoRA-FT and MoELoRA experience continuous degradation right from stage 1 (ending at $-18.78$ and $-14.63$), SEFE and HiDe-LLaVA accumulate interference as the sequence lengthens, ultimately falling to $-11.33$ and $-9.20$. In comparison, SAME remains near zero across early stages but drops to $-2.41$ at stage 5 due to accumulated parameter interference over the extended stream. In sharp contrast, \method maintains a remarkably stable, near-zero retention curve across all sequential stages, culminating at $\text{BWT}=-0.10$. This demonstrates that decoupling shared generic capabilities and enforcing orthogonal private subspace isolation effectively prevents destructive parameter interference regardless of sequence depth or cross-domain similarity.

\subsection{Further Analysis}
\label{sec:further-analysis}
\noindent\textbf{Module ablations.}
\label{sec:ablations}
We ablate the three core design components of \method on UCIT (Table~\ref{tab:module-ablation}): (1)~\emph{w/o Dynamic role allocation} enforces static assignment; (2)~\emph{w/o Activation-weighted merge} replaces activation weighting with naive equal averaging; and (3)~\emph{w/o Multi-key retention} collapses each generic slot's source keys into a single mean vector. As shown in Table~\ref{tab:module-ablation}, removing dynamic role allocation incurs the sharpest degradation in both overall accuracy ($-1.51$ points) and retention (BWT drops from $-0.10$ to $-0.52$), highlighting the importance of adaptive compatibility partitioning for reducing inter-task interference. Disabling activation-weighted merging leads to a $1.16$-point drop in overall accuracy, indicating that activation-aware weighting contributes to preserving task-relevant functional information, particularly for tasks such as CLEVR. Meanwhile, eliminating multi-key retention degrades accuracy to $72.95$ and worsens backward transfer to $-0.45$, supporting the benefit of retaining individual source keys for more reliable historical expert retrieval.

\begin{figure}[t]
\centering
\includegraphics[width=\linewidth]{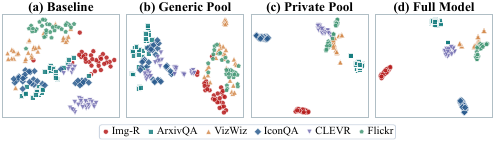}
\vspace{-4mm}
\caption{\textbf{t-SNE visualization on UCIT (Layer~16).} 
(a)~Baseline suffers from severe representational disorder across tasks. 
(b)~Generic Pool forms a continuous shared manifold across tasks without task segregation. 
(c)~Private Pool isolates task-specific features into well-separated, directional clusters.
(d)~Full \method model combines both pools into well-structured, discriminative task representations.}
\label{fig:tsne-manifold}
\vspace{-4mm}
\end{figure}

\looseness=-1
\noindent\textbf{Representation visualization.}
\label{sec:tsne-disentanglement}
Figure~\ref{fig:tsne-manifold} visualizes t-SNE projections of Layer~16 representations on UCIT to examine how the dual pools organize latent features.
As shown in Figure~\ref{fig:tsne-manifold}(a), representations from the MoE baseline (MoELoRA) suffer from severe feature entanglement across tasks due to unconstrained routing interference.
In contrast, features from the Generic Pool branch (Figure~\ref{fig:tsne-manifold}(b)) form a cohesive, continuous manifold without task segregation, confirming that consolidated generic operators capture shared cross-task priors.
Meanwhile, the Private Pool branch (Figure~\ref{fig:tsne-manifold}(c)) projects task-specific features into well-separated, directional clusters, verifying that private subspace preservation and routing regularization effectively decouple specialized representations.
Finally, the full \method model (Figure~\ref{fig:tsne-manifold}(d)) harmonizes both branches into compact, well-discriminated task representations that balance shared structural coherence with task specialization.

\begin{figure}[t]
\centering
\includegraphics[width=\linewidth]{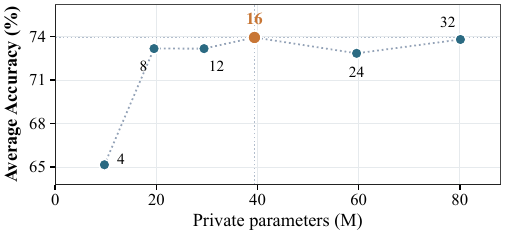}
\vspace{-6mm}
\caption{Private capacity on UCIT. Labels denote rank budgets; orange highlights the unchanged default rank 16. Each point is a full task-sequence run. Counts include Private factors only.}
\label{fig:capacity-routing}
\vspace{-4mm}
\end{figure}

\noindent\textbf{Private rank sensitivity.}
We determine the default private basis rank $R=16$ on a held-out validation split of UCIT, and Figure~\ref{fig:capacity-routing} illustrates the performance sensitivity across rank budgets $\{4, 8, 12, 16, 24, 32\}$. As observed, an overly restricted rank budget ($R=4$) degrades performance due to insufficient representational capacity. Increasing the rank to 8 and 12 rapidly restores accuracy to competitive levels, while scaling beyond 16 (e.g., $R=24, 32$) yields fluctuating or marginal performance gains at the expense of substantially higher parameter footprint. Therefore, $R=16$ provides a practical balance between task accuracy and parameter storage.
\begin{figure}[t]
\centering
\includegraphics[width=\linewidth]{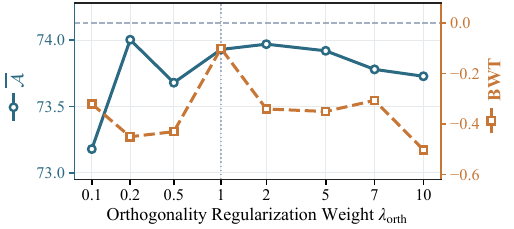}
\caption{Sensitivity to routing orthogonality regularization weight $\lambda_{\rm orth}$ on UCIT in terms of final average score $\overline{\mathcal{A}}$ and backward transfer BWT (default $\lambda_{\rm orth}=1.0$).}
\label{fig:orth-sensitivity}
\end{figure}

\noindent\textbf{Routing orthogonality sensitivity.}
\label{sec:orth-sensitivity}
We select the default weight $\lambda_{\rm orth}=1.0$ on the held-out UCIT validation split, and Figure~\ref{fig:orth-sensitivity} evaluates sensitivity across $\lambda_{\rm orth} \in [0.1, 10.0]$. Across all tested weights, the final average score $\overline{\mathcal{A}}$ remains remarkably stable, showing strong robustness. For knowledge retention, moderate regularization ($\lambda_{\rm orth}=1.0$) yields the best backward transfer by preventing routing interference between new and historical routing pairs. Conversely, excessively strong penalties over-constrain key-space exploration and slightly degrade retention. Thus, $\lambda_{\rm orth}=1.0$ strikes an effective balance between routing isolation and downstream task adaptation.

\begin{table}[t]
\centering
\caption{Task-order sensitivity on UCIT. R, A, V, I, C, F denote ImageNet-R, ArxivQA, VizWiz, IconQA, CLEVR, and Flickr30k.}
\label{tab:task-order}
\small
\setlength{\tabcolsep}{2pt}
\begin{tabular*}{\linewidth}{@{\extracolsep{\fill}}lcccc@{}}
\toprule
Order & RAVICF & AIRFCV & IFRCAV & FCIVAR\\
\midrule
$\overline{\mathcal{A}}\uparrow$ & {73.93} & 73.39 & 71.83 & 72.90\\
BWT$\uparrow$ & $-0.10$ & $0.85$ & $0.07$ & {$1.70$}\\
\bottomrule
\end{tabular*}
\vspace{-4mm}
\end{table}

\looseness=-1
\noindent\textbf{Task-order robustness.}
To evaluate robustness to task sequences, we compare the default order RAVICF against three permutations on UCIT: AIRFCV, IFRCAV, and the reverse order FCIVAR (Table~\ref{tab:task-order}). Across all four sequences, final average scores remain consistent within a narrow range from $71.83$ (IFRCAV) to $73.93$ (RAVICF), with AIRFCV closely matching at $73.39$ and FCIVAR reaching $72.90$. Meanwhile, backward transfer stays near zero or positive across every order ($\text{BWT}\in[-0.10, 1.70]$), reaching $0.85$ on AIRFCV and $1.70$ on FCIVAR. Their positive BWT values coexist with slightly lower final means than RAVICF, indicating that acquisition strength varies modestly with task position while continual retention remains consistently stable without catastrophic forgetting.

\begin{table}[t]
\centering
\caption{Efficiency and retention comparison under identical hardware. Trainable counts adaptation parameters; Train denotes cumulative training hours; Latency and Throughput measure amortized per-sample time and output token rate; $\overline{\mathcal{A}}$ and BWT denote final average score and backward transfer.}
\label{tab:efficiency}
\footnotesize
\setlength{\tabcolsep}{1.8pt}
\begin{tabular}{lcccccc}
\toprule
\raisebox{0.8ex}{Method} & \shortstack{Trainable\\(M)$\downarrow$} & \shortstack{Train\\(h)$\downarrow$} & \shortstack{Latency\\(ms/sample)$\downarrow$} & \shortstack{Throughput\\(Tokens/s)$\uparrow$} & \shortstack{$\overline{\mathcal{A}}$\\$\uparrow$} & \shortstack{BWT\\$\uparrow$} \\
\midrule
MoELoRA & 246.69 & 11.68 & 583.91 & 10.25 & 58.98 & -14.63 \\
SAME    & 73.29  & 10.42 & 404.99 & 14.30 & 67.12 & -2.41  \\
\rowcolor{tablefocus}
\method & {78.12} & {8.15}  & {326.13} & 17.34 & {73.93} & {-0.10} \\
\bottomrule
\end{tabular}
\vspace{-4mm}
\end{table}

\looseness=-1
\noindent\textbf{Efficiency comparison.}
\label{sec:paramresults}
Table~\ref{tab:efficiency} compares adaptation parameters, training time, and inference speed on UCIT alongside retention. Compared with MoELoRA (246.69\,M), \method saves 68.3\% adaptation parameters (78.12\,M) and reduces training time by 30.2\% (8.15\,h vs. 11.68\,h). At inference, by routing through compact cores with shared projections, \method cuts latency by 44.1\% and boosts decoding throughput to 17.34 tokens/s, consistently outperforming both MoELoRA and SAME. Crucially, \method achieves near-zero forgetting ($\text{BWT}=-0.10$) and the highest overall accuracy ($\overline{\mathcal{A}}=73.93$), establishing a superior trade-off across accuracy, retention, and runtime efficiency.

\section{Conclusion}
In this paper, we presented \method, a selective dual-pool framework for multimodal continual instruction tuning. Motivated by our empirical finding that continually trained experts exhibit widespread cross-task functional redundancy alongside indispensable task-specific variations, \method selectively consolidates functionally overlapping experts into a fixed-capacity generic pool while factorizing remaining task-specific experts into shared projection bases and compact cores within a private pool. Combined with frozen historical operators and orthogonal key-value routing pairs, this architecture prevents cross-task interference and catastrophic forgetting while substantially curbing parameter growth. Extensive experiments on UCIT and CoIN demonstrate that \method achieves state-of-the-art accuracy and near-zero backward transfer with compact adaptation parameters and superior runtime efficiency.

\noindent\textbf{Limitations.} While \method successfully preserves multi-task capabilities, its generated responses inevitably inherit the intrinsic visual hallucination and safety alignment behaviors of the underlying pre-trained MLLM backbone, which warrants further research.

{\small
\bibliographystyle{ieeenat_fullname}
\bibliography{main}
}
\end{document}